\documentclass[10pt,twocolumn,letterpaper]{article}

\usepackage{cvpr}              
\definecolor{cvprblue}{rgb}{0.21,0.49,0.74}
\usepackage[pagebackref,breaklinks,colorlinks,allcolors=cvprblue]{hyperref}
\usepackage{listings}
\usepackage[accsupp]{axessibility}  

\usepackage[most]{tcolorbox}
\usepackage{multirow}
\def\paperID{2799} 
\def\confName{CVPR}
\def\confYear{2026}

\title{Thinking with Frames: Generative Video Distortion Evaluation \\via Frame Reward Model}

\author{
    Yuan Wang$^{1, 2}$\footnote[1]\ \ , 
    Borui Liao$^2$, 
    Huijuan Huang$^2$\footnote[2]\ \ \footnote[3]\ \ , 
    Jinda Lu$^1$, 
    Ouxiang Li$^{1}$,\\ 
    Kuien Liu$^3$, 
    Meng Wang$^2$,
    Xiang Wang$^1$\footnote[2]\ \ \\
    \small $^1$University of Science and Technology of China, 
    $^2$Kling Team, Kuaishou Technology,  \\
   \small $^3$ Institute of Software Chinese Academy of Sciences \\
    {\tt\small wy1001@mail.ustc.edu.cn, boruiliao@gmail.com, huanghuijuan.thu@gmail.com}
}

\begin{document}
\maketitle

\renewcommand{\thefootnote}{\fnsymbol{footnote}} 
\footnotetext[1]{Work done during internship at Kling Team, Kuaishou Technology.}
\footnotetext[2]{Corresponding authors.}
\footnotetext[3]{Project Lead.}
\renewcommand{\thefootnote}{\arabic{footnote}} 

\begin{abstract}
Recent advances in video reward models and post-training strategies have improved text-to-video (T2V) generation. While these models typically assess visual quality, motion quality, and text alignment, they often overlook key structural distortions, such as abnormal object appearances and interactions, which can degrade the overall quality of the generative video.
To address this gap, we introduce REACT, a frame-level reward model designed specifically for structural distortions evaluation in generative videos. REACT assigns point-wise scores and attribution labels by reasoning over video frames, focusing on recognizing distortions. To support this, we construct a large-scale human preference dataset, annotated based on our proposed taxonomy of structural distortions, and generate additional data using a efficient Chain-of-Thought (CoT) synthesis pipeline. 
REACT is trained with a two-stage framework: (1) supervised fine-tuning with masked loss for domain knowledge injection, followed by (2) reinforcement learning with Group Relative Policy Optimization (GRPO) and pairwise rewards to enhance reasoning capability and align output scores with human preferences. During inference, a dynamic sampling mechanism is introduced to focus on frames most likely to exhibit distortion.
We also present REACT-Bench, a benchmark for generative video distortion evaluation. Experimental results demonstrate that REACT complements existing reward models in assessing structutal distortion, achieving both accurate quantitative evaluations and interpretable attribution analysis.
\end{abstract}
\vspace{-0.5cm}    
\vspace{-1mm}
\section{Introduction}
\vspace{-0.15cm}
\begin{figure*}[!t]
\centering
\includegraphics[width=\textwidth]{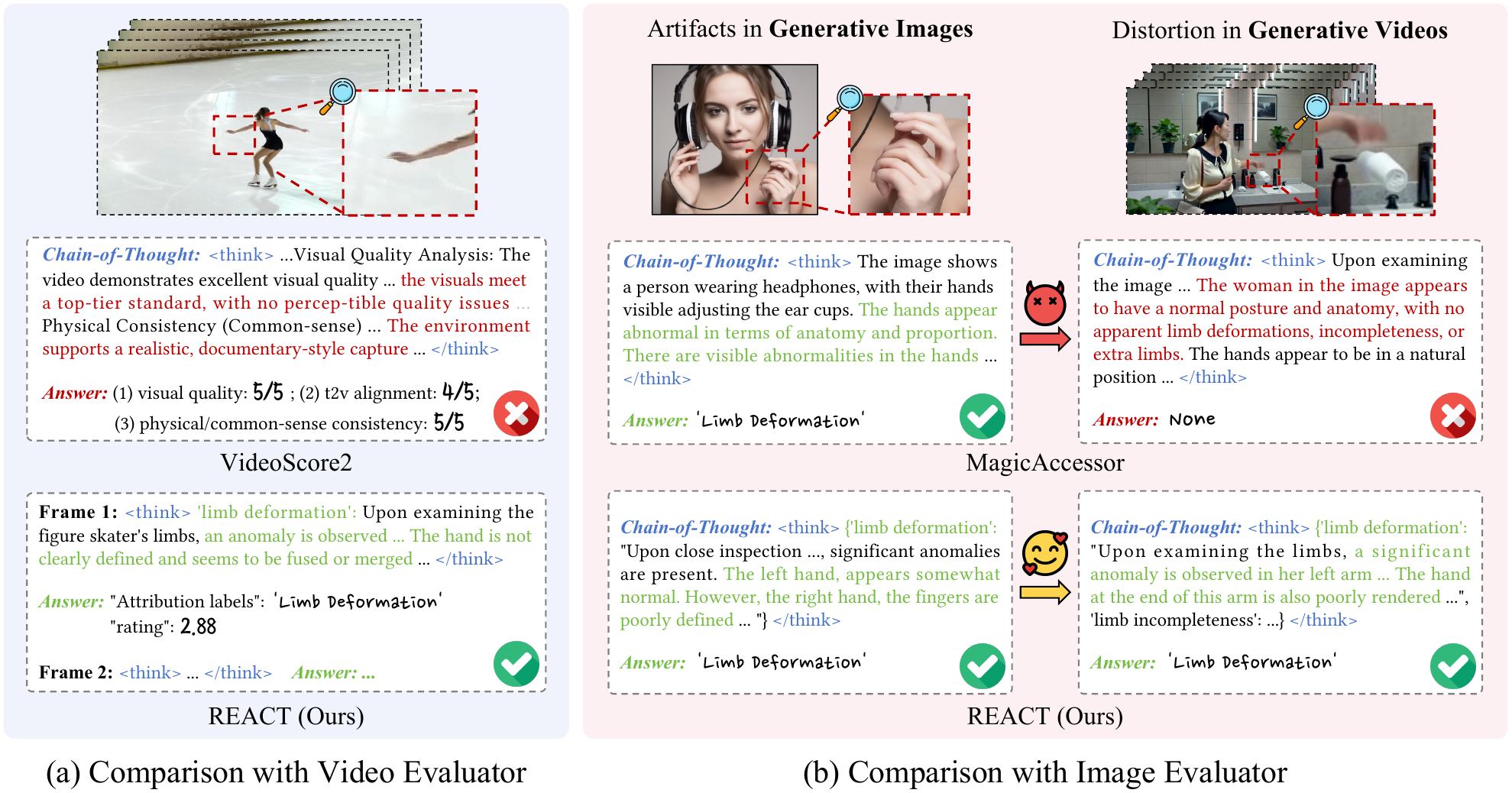}
\vspace{-7mm}
\caption{\textbf{Comparison of REACT with SOTA Video and Image Evaluators.} (a) While existing evaluators tend to assign high scores based on aesthetics and temporal consistency, even in the presence of structural defects, our REACT model outperforms them by accurately identifying structural distortions in generative videos and providing more reliable scores (b) While image evaluators excel in recognizing image artifacts, they struggle to detect distortions in generative video frames. In contrast, REACT demonstrates superior performance in recognizing and evaluating structural distortions in video frames.}
\vspace{-7mm}
\label{fig:intro}
\end{figure*}
\label{sec:intro}
Video reward models have enabled significant progress in text-to-video (T2V) generation \cite{he2024videoscore, wang2025unified, wu2025rewarddance, wang2026cookingdiffusion} by guiding models to improve visual quality, motion dynamics, and text alignment through reinforcement learning strategies \cite{vlmr1, internvl3, DeepSeekMath, liu2025flow}. However, they largely overlook \textbf{structural distortions}---abnormalities in object structures, such as abnormal object appearance (\eg, \textit{incomplete, duplicated, or deformed body parts}) or object interaction (\eg, \textit{mesh penetration, where one object unnaturally intersects with another}) in generative videos. Consequently, high scores can still be assigned to videos with severe structural distortions.


To address this limitation, we propose a \textbf{frame-level} reward model for structural distortion evaluation in generative videos, offering distinct advantages over both video-level and image-based alternatives.

\noindent\textbf{Frame-level vs. Video-level.} Compared to video-level approaches, our frame-level design is better suited for structural distortion 
assessment for three reasons: 
(1) distortions are spatially localized and detectable within individual frames; 
(2) existing video reward models operate at low sampling rates (\eg, 2 fps), limiting their ability to capture frame-specific artifacts; 
(3) frame-level annotation is significantly more efficient, enabling large-scale dataset construction from limited video samples.

\noindent\textbf{Frame-level vs. Image-based.} While image quality assessment models have explored structural distortions \cite{wang2025magicmirror, yang2025heie, ma2025evaluating, wang2024detecting, li2025improving}, they cannot be directly applied to videos due to a critical domain gap. Specifically, as illustrated in Fig.~\ref{fig:intro}, video distortions exhibit fundamentally different characteristics: unlike the sharp, well-defined artifacts in generated images, video distortions manifest as blurry, fragmented regions caused by temporal inconsistencies and motion dynamics. Such a domain gap hinders image-based evaluators from effectively capturing video-specific distortions, resulting in degraded performance when transferred to the videos.

Therefore, we propose \textbf{REACT} (\textbf{Re}ward model for \textbf{a}ssessing stru\textbf{ct}ural distortions), a frame-level model 
that provides both point-wise scores and interpretable attribution labels for structural distortions. 
Inspired by Chain-of-Thought (CoT) reasoning in both large language models (LLMs) \cite{DeepSeekR1} and multi-modal LLMs (MLLMs) \cite{zhang2025improve, dong2025insight}, REACT performs reasoning over video frames and conducts fine-grained analysis to identify structural distortions. Specifically, it is developed through two key components:

\noindent\textbf{Training data construction.}
We first develop a detailed taxonomy of structural distortions, allowing for a thorough analysis of these issues in current generative videos. Then a large-scale annotated dataset is collected with human preference pairs and multiple distortion categories derived from advanced T2V models. Given the limited ability of current MLLMs to capture visual cues related to structural distortion, sufficient CoT data is essential for fine-tuning. However, manually generating such CoT data is both costly and inefficient, as it requires detailed textual descriptions for each distortion type. We thus propose an efficient CoT synthesis pipeline, leveraging a grounded annotation task and advanced closed-source models Gemini-2.5-Pro \cite{Gemini-2.5-pro} to generate sufficient CoT data at a reduced cost.

\noindent\textbf{Two-Stage Training Framework.}
With this data foundation, We train REACT based on Qwen2.5-VL-7B \cite{bai2025qwen2} using a two-stage framework to generate point-wise scores and attribution labels for structural distortion analysis: (1) supervised fine-tuning (SFT) for domain knowledge injection, and (2) reinforcement learning (RL) with Group Relative Policy Optimization (GRPO) \cite{DeepSeekMath, DeepSeekR1} to enhance reasoning and scoring capabilities. In the SFT stage, we introduce a masked loss mechanism that enables effective domain knowledge injection while mitigating overfitting, thereby maintaining diverse reasoning trajectories for RL rather than rote replication of training CoT samples. In the RL stage, a pair-wise reward based on BTT loss is introduced to facilitate GRPO-based fine-tuning on human preference data, allowing the model to align pair-wise preferences while preserving point-wise scoring capability.

During inference, a dynamic frame sampling mechanism is employed to adaptively select frames most likely to exhibit distortions, enabling flexible analysis of fixed frame sampling constraints. Finally, we introduce REACT-Bench, a human preference benchmark specifically designed for structural distortion evaluation in generative videos, thereby complementing the generative video evaluation system.
Our contributions are summarized below:
\begin{itemize}
    \item A large-scale annotated dataset with a detailed taxonomy of structural distortions in generative videos, accompanied by an efficient CoT synthesis pipeline that generates additional training data to enhance model's reasoning capacity on distortion patterns.
    \item A frame-level reward model, REACT, for structural distortion evaluation in generative videos, providing both point-wise scores and detailed attribution labels.
    \item A human preference benchmark, REACT-Bench, specifically designed for structural distortion evaluation in generative videos. Extensive experiments on this benchmark demonstrate that REACT complements existing reward models by achieving accurate point-wise evaluations and interpretable attribution analysis.
\end{itemize}

\vspace{-3mm}
\section{Related Work}
\vspace{-2mm}
\label{sec:related work}
\noindent \textbf{Reward Model for Generative Video.}
With the development of the generative model \cite{Yao2025navmorph, xu2025drc, xu2025personalized, xu2025personalized_v2, xu2024diffusion, qiu2025accelerating, qiu2025accelerating_v2, wang2025precise, li2026speed, li2026easier, kling2025, hailuo2025}, reward modeling has become a key technique for aligning generative models with human preferences. In text-to-video generation, models like T2VQA \cite{kou2024subjective} and VideoScore \cite{he2024videoscore} assess video quality by directly training on human-annotated ratings, while another approach VideoReward \cite{liu2025improving} trains reward models based on human preference data using BTT loss \cite{bradley1952rank, rao1967ties}. To enhance reward performance and provide a more detailed reasoning process, \cite{wang2025unified2, wang2024lift, he2025videoscore2, wu2025rewarddance, wang2025unified} attempt to enable reward models to reason through CoT. 
However, these methods largely overlook structural distortions in generated videos, leading to unreliable evaluations.
Similarly, several works \cite{wu2025visualquality, li2025q, wu2023q, zhu2024adaptive} focus on image quality evaluation but fail to address structural distortion specifically. Although \cite{wang2025magicmirror, yang2025heie, ma2025evaluating, wang2024detecting, liu2025bridging} propose evaluators for detecting generative image artifacts, there exists a domain gap between the structural distortions in generative videos and the artifacts in generative images. This motivates us to propose a reward model specifically for evaluating structural distortions in generative videos, further complementing the video reward system.

\noindent \textbf{Reinforcement Learning.}
The integration of reinforcement learning (RL) into Large Language Models (LLMs) and Multi-modal LLMs (MLLMs) \cite{hurst2024gpt, hui2024qwen2, jaech2024openai, GPT-5, Gemini-2.5-pro} has significantly advanced their reasoning capabilities \cite{r1_reward, vlmr1, internvl3, llavar1}. This improvement arises from the shift away from models merely replicating training data during fine-tuning, to a more dynamic approach in which models refine their reasoning trajectories and enhance output quality through reward optimization. Practically, this paradigm is initially implemented using Proximal Policy Optimization (PPO) \cite{ppo, lu2025adavipaligningmultimodalllms}, an extension of the classic policy gradient algorithm. A notable breakthrough comes with the introduction of Group Relative Policy Optimization (GRPO) \cite{DeepSeekMath, DeepSeekR1}, which simplifies the calculation of advantages. GRPO has since been successfully applied to a variety of downstream tasks in visual understanding\citep{yu2024rlaif, zhou2024aligning, liu2025visual, sun2024aligning, yu2024rlhf, lu2025dama, liu2025diversegrpo, li2026enhancing}, improving the model’s ability to perform long-chain reasoning. More recently, GRPO has been also incorporated into reward modeling for visual generation tasks \cite{wu2025visualquality, he2025videoscore2, li2025q}. Building on this, we adopt the same paradigm to enhance the performance of our proposed frame-level reward model, enabling it to reason over individual frames and conduct detailed analyses of structural distortions.

\vspace{-3mm}
\section{Method}
\vspace{-1mm}
\begin{figure*}[!t]
\centering
\includegraphics[width=\textwidth]{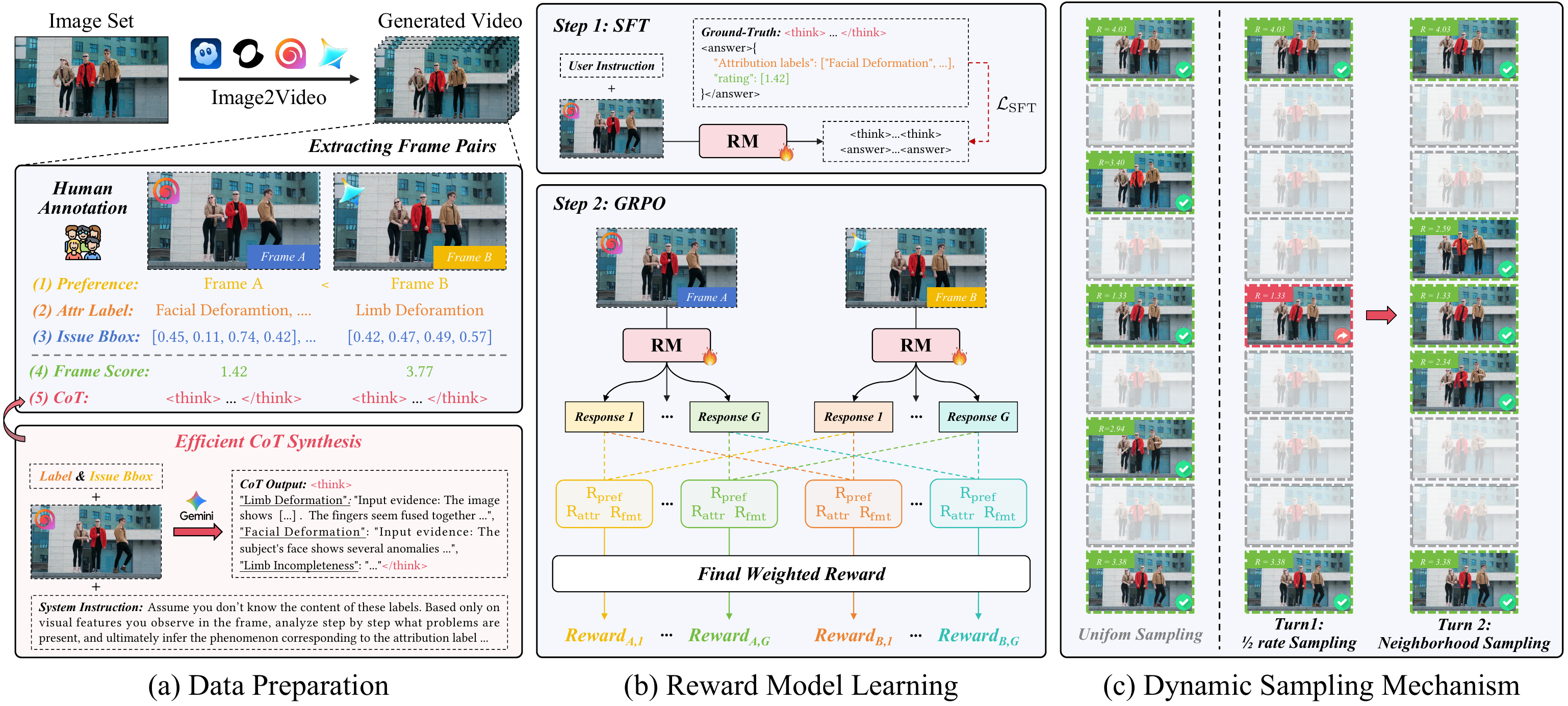}
\vspace{-6.5mm}
\caption{\textbf{Overview of REACT: Frame-Level Reward Model for Structural Distortion Evaluation.} (a) We first construct a large-scale annotated dataset, including \textcolor[HTML]{efb806}{\textbf{\textit{human preference}}} and \textcolor[HTML]{ed822e}{\textbf{\textit{attribution labels}}}, based on our proposed detailed taxonomy of structural distortions. Furthermore, we synthesize \textcolor[HTML]{e44c5e}{\textbf{\textit{CoT data}}} through an efficient pipeline that leverages human-annotated \textcolor[HTML]{4774c9}{\textbf{\textit{issue bounding boxes}}} and label-aware sampled \textcolor[HTML]{79bd48}{\textbf{\textit{frame-level scores}}}. (b) We then train REACT based on Qwen2.5-VL-7B using a two-stage training framework. During \textbf{\textit{SFT stage}}, a masked loss is applied to improve domain knowledge injection. During \textbf{\textit{GRPO stage}}, pair-wise rewards are introduced to align the output point-wise scores of REACT with human preferences. (3) Finally, frames most likely to exhibit distortions are adaptively selected with a dynamic sampling mechanism, enabling flexible analysis within fixed frame sampling constraints.}
\vspace{-6.5mm}
\label{fig:method}
\end{figure*}
\subsection{Data Preparation}
\vspace{-1.5mm}
\label{subsec:data_prepara}
\textbf{Taxonomy of Structural Distortion.}
Although existing video reward models may implicitly account for distortion within visual or motion quality evaluations, they lack a systematic analysis and taxonomy of structural distortions. To enable fine-grained assessment, we establish a detailed taxonomy that categorizes structural distortions in generative videos into two primary aspects: \textbf{abnormal object appearance} and \textbf{abnormal object interaction}. 

Abnormal object appearance describes deviations in the shape or structure of objects in generative videos. This category is further divided into animal-related and non-animal distortions. Non-animal distortions refer to abnormalities in inanimate objects such as plates and background elements. For animal-related distortions, we analyze three body parts (\ie limbs, torso, and face) and three typical distortion types: deformation, incompleteness (missing parts), and duplication (extra parts). Since incompleteness and duplication rarely occur in the torso or face, they are only considered for limbs. As a result, we define five specific categories for abnormal object appearance: limb deformation, extra limbs, limb incompleteness, torso deformation, and facial deformation. In addition, motion blur is included as it is a common artifact in video generation. Abnormal object interaction, on the other hand, refers to violations of physical plausibility in spatial relationships among objects. The primary case considered is mesh penetration, where object boundaries interpenetrate or fuse in unrealistic ways, breaking the impenetrability principle of solid matter. In summary, the proposed taxonomy covers eight distinct categories: \textbf{limb deformation}, \textbf{extra limbs}, \textbf{limb incompleteness}, \textbf{torso deformation}, \textbf{facial deformation}, \textbf{non-animal collapse and distortion}, \textbf{motion blur}, and \textbf{mesh penetration}. All collected data are annotated and compared according to these categories, with detailed definitions and visual examples provided in Appendix~\ref{taxonomy}.

\noindent \textbf{Data Collection.}
To construct the training dataset, we first collect real-world videos featuring complex motions from social media platforms. These videos are then captioned to create text prompts for generation, as the complexity of motion patterns makes it difficult for current T2V models to produce high-quality results, often leading to structural distortions. Several state-of-the-art T2V models, including Kling \cite{kling2025}, HaiLuo \cite{hailuo2025}, Seedream \cite{seedream2025}, Pika \cite{pika2023}, Sora \cite{sora2024}, and Luma \cite{luma2024}, are employed to generate videos based on these prompts. For constructing frame-level preference pairs, we use two different generation models to synthesize videos from the same prompt, pairing frames corresponding to identical timestamps. To contain some pairs share the same semantic content while differing only in visual quality, we also incorporate image-to-video (I2V) generation paradigms. Specifically, frames sampled from real videos are used as visual references to guide I2V generation, resulting in a dataset that combines outputs from both T2V and I2V models. In total, we construct over 15k pairs (\ie, approximately 30k frames) for model training.


\noindent \textbf{Efficient Chain-of-Thought Synthesis.}
To enable the MLLMs (\eg Qwen2.5-VL-7B) to reason about structural distortions in generative video frames, we construct high-quality Chain-of-Thought (CoT) data that combine attribution labels, point-wise scores, and reasoning traces
Manually creating such data is costly, as it requires detailed textual descriptions for each distortion type. This difficulty is further compounded by the limited capability of current multimodal large language models (MLLMs) to fully capture visual cues related to structural distortion, making large-scale data necessary to teach both reasoning skills and domain-specific knowledge. 

To address these challenges, we propose an efficient CoT synthesis pipeline that reformulates annotation as a grounding task. Annotators only need to draw bounding boxes around distorted regions, thereby greatly reducing annotation effort and improving quality control. Given the annotated frames and corresponding distortion regions, Gemini 2.5 Pro \cite{Gemini-2.5-pro} is prompted to simulate the reasoning process that produces the correct attribution labels and localization results, using the prompt templates described in Appendix~\ref{prompt}. The generated CoT samples are filtered based on the accuracy of their predicted labels and regions. The resulting samples are filtered by label and region accuracy, yielding 6K high-quality CoT instances for training. Since our dataset is based on frame preference pairs rather than point-wise scores, we further introduce pseudo point-wise scores for numerical supervision. For each CoT sample, a score with two decimal places is randomly assigned based on the number of distortion labels: a score in the range of [4.0, 5.0] for distortion-free frames, [3.0, 4.0] for one label, [2.0, 3.0] for two labels, and [1.0, 2.0] for three or more. Though approximate, these scores maintain human ranking consistency and promote score diversity during fine-tuning, while GRPO further aligns quantitative judgment.

\noindent \textbf{Human Annotation.} Each frame pair is annotated with human preference labels and attribution labels specifying the types of distortion. A team of 34 professional image and video evaluation experts, consisting of 20 annotators and 14 reviewers, is responsible for the annotation process. Initially, 2,000 cases are selected for annotator training, aiming for annotation accuracy above 90$\%$. The formal annotation process includes two rounds of review, with any errors in each round returned for correction. Additionally, a random sample of 10$\%$ of the annotations undergoes final quality control, achieving bounding box accuracy above 95$\%$ and attribution label accuracy above 90$\%$. This process results in 15K frame pairs with attribution labels and human preference annotations. The detailed annotation protocol is provided in Appendix~\ref{annotation}.

\vspace{-2mm}
\subsection{Reward Model Learning}
\vspace{-2mm}
\label{subsec:training_framework}
Our frame-level reward model REACT adopts Qwen2.5-VL-7B as the base model and follows a two-stage training paradigm. Specifically, we first perform supervised fine-tuning (SFT) on the CoT data to inject domain knowledge and enable the model to recognize structural distortions. Then, Group Relative Policy Optimization (GRPO) is applied to further enhance the model’s reasoning ability and encourage it to generate more accurate attribution labels and point-wise scores.

\noindent \textbf{Supervised Fine-Tuning.}
~In this stage, our goal is not only to enable the general MLLM to reason over video frames but also to accurately identify structural distortions and produce the corresponding attribution labels and point-wise scores. However, during supervised fine-tuning (SFT), excessive training iterations often lead to performance degradation in GRPO, as the model tends to overfit the training data and merely imitate the constructed CoT patterns, thereby reducing the diversity of its reasoning trajectories. At the same time, limited training steps are insufficient for effective domain knowledge injection.

To balance these objectives, we introduce a masked supervised fine-tuning strategy. Specifically, we first fine-tune the base model on the complete CoT data, where the reasoning process, attribution labels, and point-wise scores are all visible to teach it how to infer distortion patterns. Then, to prevent the model from overfitting to the reasoning traces, we perform masked SFT, where only the final attribution labels and scores are used for loss computation. This approach refines the accuracy of labeling and scoring while avoiding excessive reliance on predefined reasoning paths.

\noindent \textbf{Reinforcement Learning via GRPO.}
~To strengthen the model’s reasoning process---thereby improving its ability to detect structural distortions and generate accurate point-wise scores---we employ GRPO to refine the policy through group-wise relative comparisons of alternative reasoning trajectories.

Given a text prompt $\boldsymbol{c}$ and a video frame $\boldsymbol{f}$, the objective is to fine-tune our REACT model to generate a point-wise score in the range of $[1, 5]$ and corresponding attribution labels through step-by-step reasoning guided by the prompt, as shown in Fig.~\ref{box:generation_instruction}.
The standard GRPO samples a group of responses $\{\boldsymbol{o}_1, \boldsymbol{o}_2, \dots, \boldsymbol{o}_G\}$ based on input $\boldsymbol{q}=\{\boldsymbol{c}, \boldsymbol{f}\}$ from the old policy model $\pi_{\theta_{\text{old}}}$, with rollout size $G$. The advantage of the $i$-th is computed by normalizing the rewards among the group. GRPO updates the policy model $\pi_{\theta}$ using a clipped objective, along with a KL penalty term, formulated as:
\begin{align}
A_{i}
= \frac{R(\boldsymbol{o}_i) - mean(\{ R(\boldsymbol{o}_1), R(\boldsymbol{o}_2), \dots, R(\boldsymbol{o}_G)\})}{std(\{ R(\boldsymbol{o}_1), R(\boldsymbol{o}_2), \dots, R(\boldsymbol{o}_G)\})},
\end{align}
\begin{align}
& \mathcal{J}_{\mathrm{GRPO}}(\theta) = \mathbb{E}_{\boldsymbol{q} \sim \mathcal{Q}, \{\boldsymbol{o}_i\}_{i=1}^G \sim \pi_{\theta_{\text{old}}}(\boldsymbol{o} \mid \boldsymbol{q})} \Bigg\{ \frac{1}{G} \sum_{i=1}^G \frac{1}{|\boldsymbol{o}_i|} \sum_{t=1}^{|\boldsymbol{o}_i|} \nonumber \\
& - \beta \mathbb{D}_{KL} (\pi_{\theta} \| \pi_{\text{ref}})+ \min \left[ \frac{\pi_{\theta}(\boldsymbol{o}_{i,t} \mid \boldsymbol{q}, \boldsymbol{o}_{i,<t})}{\pi_{\theta_{\text{old}}}(\boldsymbol{o}_{i,t} \mid \boldsymbol{q}, \boldsymbol{o}_{i,<t})} A_{i,t}, \right. \nonumber \\
& \left. \mathrm{clip}\left( \frac{\pi_{\theta} (\boldsymbol{o}_{i,t} \mid \boldsymbol{q}, \boldsymbol{o}_{i,<t})}{\pi_{\theta_{\text{old}}}(\boldsymbol{o}_{i,t} \mid \boldsymbol{q}, \boldsymbol{o}_{i,<t})}, 1-\epsilon, 1+\epsilon \right) A_{i,t} \right] \Bigg\}. 
\end{align}
Here, $r_i$ refers to the reward of the $i$-th response $o_{i}$, $\epsilon$ controls the clipping range of the importance sampling ratio, and $\beta$ is the penalty strength for how much the current policy $\pi_\theta$ deviates from the reference policy $\pi_{ref}$.

Although our training dataset includes human preference pairs and attribution labels, the absence of point-wise scores prevents us from directly calculating rewards based on the difference between predicted and ground-truth scores for advantage estimation in GRPO. To address this, we propose a \textbf{pairwise reward} based on the BTT loss \cite{rao1967ties}, which allocates a reward to each rollout within a group by calculating pair-wise scores based on the training frame pairs. 
Specifically, given a frame pair $\{\boldsymbol{f}^{A}, \boldsymbol{f}^{B}\}$ sampled from the training dataset, REACT generates rollouts for each frame separately, prompted by text prompt $c$, resulting in two groups: $\{\boldsymbol{o}_1^{A}, \boldsymbol{o}_2^{A}, \dots, \boldsymbol{o}_G^{A}\}$ and $\{\boldsymbol{o}_1^{B}, \boldsymbol{o}_2^{B}, \dots, \boldsymbol{o}_G^{B}\}$. The reward for each rollout $\boldsymbol{o}_i^{j}$ (where $j=A~\text{or}~B$) consists of three components: format reward, attribution accuracy reward, and preference reward.
\begin{itemize}
    \item \textit{Format Reward.}~To ensure that the output follows the format specified in the text prompts, we assign a format reward $R_{\text{fmt}}(\boldsymbol{o}_{i}^{j})$ of 1 if the reasoning process is contained within \texttt{<think></think>} and the attribution labels and point-wise score are within \texttt{<answer></answer>}. Otherwise, the format reward is set to 0.
    \item \textit{Attribution Accuracy Reward.} Since each frame is annotated with detailed distortion issues, the attribution accuracy reward $R_{\text{attr}}$ is calculated by comparing the output attribution labels with the ground truth. Specifically:
    \begin{equation}
        R_{\text{attr}}(\boldsymbol{o}_{i}^{j}) = 0.6\cdot a_{\text{right}}-0.2\cdot (a_{\text{wrong}}+a_{\text{missing}}),
    \end{equation}
    where $a_{\text{right}}$, $a_{\text{wrong}}$, $a_{\text{miss}}$ refer to the right, wrong, and missing attribution labels in the $\boldsymbol{o}_{i}^{j}$, respectively.
    \item \textit{Preference Reward.} To allocate the preference reward for each rollout of each frame within the pair, we calculate the probabilities of each possible preference, rather than directly comparing the predicted scores and using binary rewards based on ground truth. Inspired by \cite{liu2025improving}, we compute the preference probabilities as follows:
    \begin{align}
       & P(\boldsymbol{o}_{i}^{A} \succ \boldsymbol{o}_{i}^{B}|\boldsymbol{c}) = \frac{e^{s_{i}^{A}}}{\theta e^{s_{i}^{A}}+e^{s_{i}^{B}}}, \\
        &P(\boldsymbol{o}_{i}^{A} \prec \boldsymbol{o}_{i}^{B}|c) = \frac{e^{s_{i}^{B}}}{\theta e^{s_{i}^{A}}+e^{s_{i}^{B}}}, \\
        &P(\boldsymbol{o}_{i}^{A}=\boldsymbol{o}_{i}^{B}|c) = \frac{(\theta^2-1)e^{s_{i}^{A}}e^{s_{i}^{B}}}{(e^{s_{i}^{A}}+\theta e^{s_{i}^{B}})(\theta e^{s_{i}^{A}}+e^{s_{i}^{B}})}.
    \end{align}
    Here, \( s_i^A \) and \( s_i^B \) are the point-wise scores of frames $A$ and $B$, respectively, extracted  from $\boldsymbol{o}_{i}^{A}$ and $\boldsymbol{o}_{i}^{B}$, as predicted by REACT. The preference reward is computed as:
\begin{align}
    R_{\text{pref}}(\boldsymbol{o}_{i}^{A}, \boldsymbol{o}_{i}^{B})&=\mathbb{I}(\boldsymbol{f}^{A}\succ \boldsymbol{f}^{B})\text{log}P(\boldsymbol{o}_{i}^{A}\succ \boldsymbol{o}_{i}^{B}|\boldsymbol{c})\nonumber\\
    &+\mathbb{I}(\boldsymbol{f}^{A}\prec \boldsymbol{f}^{B})\text{log}P(\boldsymbol{o}_{i}^{A}\prec \boldsymbol{o}_{i}^{B}|\boldsymbol{c})\nonumber\\
    &+\mathbb{I}(\boldsymbol{f}^{A}=\boldsymbol{f}^{B})\text{log}P(\boldsymbol{o}_{i}^{A}=\boldsymbol{o}_{i}^{B}|\boldsymbol{c}),
\end{align}
where $\mathbb{I}(\cdot)$ is an indicator function that equals 1 when the ground truth preference is satisfied, and 0 otherwise. The hyper-parameter $\theta$ controls the tendency towards ties, and we set it to 5, following \cite{liu2025improving}.
\end{itemize}
The final reward for each rollout is computed as follows:
\begin{equation}
R(\boldsymbol{o}_i^j) = \lambda_1 R_{\text{fmt}}(\boldsymbol{o}_i^j) + \lambda_2 R_{\text{attr}}(\boldsymbol{o}_i^j) + \lambda_3 R_{\text{pref}}(\boldsymbol{o}_i^A, \boldsymbol{o}_{i}^{B}),
\end{equation}
where the $\lambda_{1}$, $\lambda_{2}$ and $\lambda_{3}$ are the weights assigned to each reward component.

\vspace{-1mm}
\subsection{Dynamic Sampling Mechanism}
\vspace{-1mm}
Existing video-level reward sampling typically selects frames at fixed intervals determined by the sampling frame rate (fps). However, when the sampling fps is low relative to the video fps, this strategy risks missing critical distorted frames. Moreover, generative videos often exhibit strong temporal consistency, suggesting that distortion patterns in adjacent frames are likely correlated.
Therefore, we introduce a dynamic sampling mechanism that operates in two stages. In the first stage, frames are sampled at half the fps and analyzed using the REACT model. Based on the score distribution, three cases can be categorized into the following cases:
\begin{itemize}
    \item If all the sampled frames have high scores, exceeding a high threshold, they are likely distortion-free, and the remaining frames are sampled farther apart in the second stage,  where frames between those selected in the first stage are sampled.
    \item If the scores fall below a low threshold, it indicates structural distortions, prompting us to sample adjacent frames within a 1/4 fps interval from those selected in the first stage.
     \item If neither of the above two cases occurs, it indicates a mix of distortion-free and distorted frames. In this case, we prioritize frames with scores lower than the mean and sample two frames randomly within a 1/4 fps interval around these low-score frames.
\end{itemize}
Finally, the overall video score is computed by averaging the scores from both the first and second stages of sampling. This dynamic sampling mechanism enhances the probability of selecting problematic frames while maintaining a fixed sampling count.

\vspace{-2mm}
\section{Experiments}
\vspace{-1mm}
\subsection{Experimental Setups.}

\noindent \textbf{Implementation.}
We adopt Qwen2.5-VL-7B as the base model for REACT. During the supervised fine-tuning (SFT) stage, the model is trained on the constructed Chain-of-Thought (CoT) dataset, with a learning rate of 5e-4, and LoRA applied for fine-tuning with a rank of 32. In the first epoch, the full responses are used for loss computation, while in the second epoch, the reasoning trajectories are masked to prevent overfitting to explicit reasoning patterns. We employ the AdamW optimizer with a weight decay of 0.01 and a batch size of 64 during SFT. In the reinforcement learning (RL) stage, we apply Group Relative Policy Optimization (GRPO) with a learning rate of 1.0e-6 and a rollout group size of $G=8$, using the same optimizer configuration as in SFT. GRPO training is conducted for 300 steps, with a rollout batch size of 256 and an update mini-batch size of 64. During inference, a dynamic frame sampling strategy is employed at 2 fps per video, and all results are evaluated on the REACT-Bench benchmark.  

\noindent \textbf{Baseline.}
For the human preference alignment task, \ie, ranking video quality based on the severity of structural distortions, we compare our REACT with several state-of-the-art (SOTA) video reward models, including VideoReward \cite{liu2025improving}, VideoScore2 \cite{he2025videoscore2}, and UnifiedReward \cite{wang2025unified}. In addition, image-based reward models such as Q-Insight \cite{dong2025insight} and VisualQuality-R1 \cite{wu2025visualquality} are also included for comparison by evaluating video quality at the frame level, consistent with the evaluation setting of our REACT. For the distortion recognition task, \ie, determining whether a video frame exhibits structural distortions, we adopt MagicAssessor \cite{wang2025magicmirror}, a SOTA image evaluator for generative artifacts, as the baseline. Furthermore, we include several general multimodal large language models (MLLMs) for comprehensive comparison. Specifically, Gemini-2.5-Pro \cite{Gemini-2.5-pro}, Gemini-2.5-Flash \cite{Gemini-2.5-flash}, and Qwen2.5-VL-7B \cite{bai2025qwen2} are evaluated on both two tasks, while GPT-4o \cite{hurst2024gpt} and GPT-o3 \cite{gpt_o3_openai} are used exclusively for the distortion recognition task. In addition, we further evaluate the effectiveness of our reward model in improving generated video quality on a text-to-video generation benchmark, \ie VBench \cite{huang2024vbench}, with results reported in Appendix \ref{add_exp: rl}.

\noindent \textbf{REACT-Bench}
To comprehensively evaluate our REACT model on both human preference alignment and structural distortion recognition, we construct a new benchmark named REACT-Bench, consisting of two complementary subsets: REACT-Video and REACT-Frame.
REACT-Video comprises 500 human-annotated video pairs, each labeled with pairwise preference scores reflecting the quality differences related to distortion between two generated videos. The annotation follows the criteria described in Section~\ref{subsec:data_prepara}. REACT-Frame contains 2.1K annotated video frames and serves as a fine-grained sub-benchmark dedicated to frame-level distortion recognition. Each frame is annotated with detailed attribution labels aligned with our structural distortion taxonomy, covering both distorted and normal cases. Together, these two subsets establish a comprehensive evaluation framework for assessing both preference alignment and structural distortion understanding, providing a complementary benchmark for future research in reward modeling for generative video quality assessment. 

\vspace{-2mm}
\subsection{Main Results}
\label{exp:main_results}
\vspace{-1mm}
\noindent \textbf{Human Preference Alignment.}
\begin{table}
  \centering
  \caption{\textbf{Comparison of REACT with SOTA Models on Human Preference Alignment.} The best and second-best results are highlighted in \textbf{bold}, and ``+Rep'' indicates that the model is evaluated with a refined prompt. Our REACT model outperforms existing methods, achieving the highest accuracy in preference assignment based on structural distortion}
  \renewcommand{\arraystretch}{1.3}  
  \vspace{-2mm}
\resizebox{\hsize}{!}{\begin{tabular}{ccccccc}
\hline
\multicolumn{1}{c|}{}                                                              & \multicolumn{3}{c|}{Acc w/ Tie}         & \multicolumn{3}{c}{Acc w/o Tie} \\ \cline{2-7} 
\multicolumn{1}{c|}{\multirow{-2}{*}{Model}}                                       & VQ    & MQ    & \multicolumn{1}{c|}{Overall} & VQ       & MQ       & Overall    \\ \hline
\multicolumn{7}{c}{Video Evaluator}                                                                                                                                  \\ \hline
\multicolumn{1}{c|}{VideoScore2}                                                   & 0.362 & 0.364 & \multicolumn{1}{c|}{0.342}   & 0.550    & 0.540    & 0.521      \\
\multicolumn{1}{c|}{UnifiedReward}                                                 & 0.390 & 0.400 & \multicolumn{1}{c|}{0.416}   & 0.707    & 0.674    & 0.701      \\
\multicolumn{1}{c|}{VideoReward}                                                   & 0.407 & 0.417 & \multicolumn{1}{c|}{0.415}   & 0.524    & 0.572    & 0.551      \\ \hline
\multicolumn{7}{c}{\textit{General Multimodal Language Model}}                                                                                                                \\ \hline
\multicolumn{1}{c|}{Qwen2.5-VL-7B}                                                 & \multicolumn{3}{c|}{0.376}                   & \multicolumn{3}{c}{0.509}        \\
\multicolumn{1}{c|}{Qwen2.5-VL-32B} & \multicolumn{3}{c|}{0.364}                   & \multicolumn{3}{c}{0.583}        \\
\multicolumn{1}{c|}{Gemini-2.5-Flash}                                              & \multicolumn{3}{c|}{0.384}                   & \multicolumn{3}{c}{0.553}        \\
\multicolumn{1}{c|}{Gemini-2.5-Pro}                                                & \multicolumn{3}{c|}{0.370}                   & \multicolumn{3}{c}{0.534}        \\ \hline
\multicolumn{7}{c}{\textit{Image Eavalutor}}                                                                                                                                  \\ \hline
\multicolumn{1}{c|}{Q-insight}                                                      & \multicolumn{3}{c|}{0.384}                   & \multicolumn{3}{c}{0.559}        \\
\multicolumn{1}{c|}{Q-insight \footnotesize{(\textbf{+Rep)}}}                                                & \multicolumn{3}{c|}{0.354}                   & \multicolumn{3}{c}{0.552}        \\
\multicolumn{1}{c|}{VisualQuality-R1}                                               & \multicolumn{3}{c|}{0.376}                   & \multicolumn{3}{c}{0.610}        \\
\multicolumn{1}{c|}{VisualQuality-R1 \footnotesize{(\textbf{+Rep})}}                                                & \multicolumn{3}{c|}{0.376}                   & \multicolumn{3}{c}{0.586}        \\ \hline
\multicolumn{7}{c}{\textit{Our REACT}}                                                                                                                                            \\ \hline
\multicolumn{1}{c|}{REACT}                                                          & \multicolumn{3}{c|}{\textbf{0.610}}                   & \multicolumn{3}{c}{\textbf{0.813}}        \\ \hline
\end{tabular}}
\vspace{-5mm}
\label{tabel_score}
\end{table}
We first evaluate the performance of REACT on human preference alignment using the REACT-Video. As shown in Table~\ref{tabel_score}, we compare REACT with state-of-the-art (SOTA) video evaluators, image evaluators, and general multimodal large language models (MLLMs). For image evaluators such as Q-insight and VisualQuality-R1, which rely on MLLMs and are sensitive to prompt design, we refine their prompts using our annotation guidelines to strengthen their ability to identify structural distortions. For general MLLMs, evaluation is performed at the video level with a sampling rate of 2 fps. For video evaluators typically assess three aspects, \ie, visual quality (VQ), motion quality (MQ), and text alignment (TA). VQ measures aesthetic attributes like resolution, clarity, and color fidelity, while MQ evaluates the smoothness and physical plausibility of movements, and TA checks the semantic consistency between the video and the input prompt. Since structural distortions are more closely related to VQ and MQ, we report their average as the overall score. Detailed settings are provided in Appendix~\ref{add_exp}.

Although UnifiedReward achieves the strongest performance among existing video evaluators, with accuracies of 0.416 (w/ tie) and 0.701 (w/o tie), it still falls notably short of REACT, which reaches 0.610 and 0.813 on the same metrics. This performance gap indicates that current video evaluators insufficiently account for structural distortion and tend to assign high scores to videos that exhibit good aesthetics or temporal consistency, even when structural defects are present. A similar pattern is observed for image evaluators and general MLLMs. Despite refining the prompts of Q-insight and VisualQuality-R1 to better emphasize structural distortion cues, their accuracies remain substantially lower than REACT (0.354–0.384 w/ tie; 0.552–0.610 w/o tie), highlighting the domain gap between distortions in generated images and those in generated videos. General MLLMs such as Gemini-2.5-Pro and Qwen2.5-VL-7B perform even worse, underscoring their limited capacity to reliably identify structural defects in video content. In contrast, REACT consistently achieves the highest accuracy across all settings, yielding a relative improvement of 20–40$\%$ over existing evaluators. These results validate the necessity of explicitly modeling structural distortion in generative video evaluation.

To further validate the effectiveness of REACT in human preference alignment, we conduct additional evaluations on benchmarks including GenAI-Bench \cite{jiang2024genai} and VideoGen-RewardBench \cite{liu2025improving}, with results provided in Appendix \ref{add_exp: align}.

\noindent \textbf{Distortion Recognition.}
\begin{table}
  \centering
  \caption{\textbf{Comparison of REACT with SOTA Models in Distortion Recognition.} The best and second-best results are marked in \textbf{bold} and \underline{underlined}, respectively. Our REACT model achieves the highest F1-score in distinguishing distorted frames, demonstrating its superior accuracy in recognizing structural distortions in video frames.}
  \renewcommand{\arraystretch}{1.3}  
  \vspace{-2mm}
\resizebox{\hsize}{!}{\begin{tabular}{ccccccc}
\hline
\multicolumn{1}{c|}{}                        & \multicolumn{3}{c|}{Distorted Frame}                                  & \multicolumn{3}{c}{Normal Frame}                                        \\ \cline{2-7} 
\multicolumn{1}{c|}{\multirow{-2}{*}{Model}} & Recall         & Precision      & \multicolumn{1}{c|}{F1-Score}             & Recall         & Precision      & F1-Score                                \\ \hline
\multicolumn{7}{c}{\textit{General Multimodal Large Language Models}}                                                                                                                                                            \\ \hline
\multicolumn{1}{c|}{Gemini-2.5-Pro}          & {\underline{0.509}}    & 0.902          & \multicolumn{1}{c|}{{\underline{0.650}}}     & 0.715          & 0.219          & 0.335                                 \\
\multicolumn{1}{c|}{Gemini-2.5-Flash}        & 0.375          & 0.919          & \multicolumn{1}{c|}{0.532}          & 0.829          & 0.204          & 0.327                                 \\
\multicolumn{1}{c|}{GPT-o3}                  & 0.485          & 0.947          & \multicolumn{1}{c|}{0.641}          & 0.859          & {\underline{ 0.243}}    & {\underline{ 0.379}}                           \\
\multicolumn{1}{c|}{GPT-4o}                  & 0.332            & 0.924              & \multicolumn{1}{c|}{0.488}              & 0.856             & 0.196              & 0.319                                     \\
\multicolumn{1}{c|}{Qwen2.5-VL-7B}           & 0.089          & {\textbf{ 0.957}}    & \multicolumn{1}{c|}{0.162}          & \textbf{0.979} & 0.172          & 0.292                                 \\
\multicolumn{1}{c|}{Qwen2.5-VL-32B}          &  0.099             & 0.935              & \multicolumn{1}{c|}{0.179}              & 0.965              & 0.171              & 0.291                                     \\ \hline
\multicolumn{7}{c}{\textit{Image Evaluator}}                                                                                                                                                            \\ \hline
\multicolumn{1}{c|}{VisualQuality-R1}        & 0.121          & \underline{0.955} & \multicolumn{1}{c|}{0.215}          & {\underline{ 0.971}}    & 0.176          & 0.297                                 \\
\multicolumn{1}{c|}{Q-insight}               & 0.204          & 0.918          & \multicolumn{1}{c|}{0.334}          & 0.906          & 0.180           & 0.300                                   \\
\multicolumn{1}{c|}{MagicAccessor}           & 0.407          & 0.867          & \multicolumn{1}{c|}{0.554}          & 0.676          & 0.180           & 0.285                                 \\ \hline
\multicolumn{7}{c}{\textit{Our REACT}}                                                                                                                                                                  \\ \hline
\multicolumn{1}{c|}{REACT}                    & \textbf{0.866} & 0.825          & \multicolumn{1}{c|}{\textbf{0.845}} & 0.594          & \textbf{0.771} & {\textbf{0.671}} \\ \hline
\end{tabular}}
\vspace{-6mm}

\label{tabel_attr}
\end{table}
To evaluate the structural distortion recognition ability of our REACT model, we compare it with current state-of-the-art (SOTA) image evaluators and general multimodal large language models (MLLMs) using our proposed REACT-Frame (i.e., frame-level sub-benchmark). Within these models, VisualQuality-R1 and Q-insight are trained to give a point-wise score, according to the quality of generative image. However, their are constructed based on MLLMs, then we designed use prompt to guided them to thinking about distortions. In the experiments, frames with distortion issues are labeled as distorted, while frames without any distortion are considered normal, and Precision, Recall and F1-Score are used to evaluate the accuracy of distortion recognition. As shown in Table~\ref{tabel_attr}, REACT outperforms existing methods in recognizing structural distortions in generative videos, achieving the highest F1-score for both distorted and normal frames. This indicates that REACT can accurately identify frames with structural distortion while maintaining high accuracy in distinguishing normal frames without falsely classifying them as distorted. In contrast, current general MLLMs and SOTA image evaluators lag behind REACT. While these models generally achieve high precision for distorted frames and high recall for normal frames, their low F1-scores indicate a tendency to classify distorted frames as normal. This highlights the difficulty that general MLLMs face in recognizing structural distortions. It also underscores the challenges that image evaluators encounter when assessing distortions in generative videos, due to the domain gap. Unlike these models, REACT demonstrates a superior ability to accurately recognize frames with structural distortion issues.

\vspace{-2mm}
\subsection{Ablation Study}
\vspace{-1mm}
\label{subsec:ablation}
\begin{table}
  \footnotesize 
  \centering
  \caption{\textbf{Ablation Study on RL Starting Point, Reward Design, and Sampling Mechanism in Human Preference Alignment.} Our REACT model with the default settings performs best.}
  \renewcommand{\arraystretch}{1.3}  
  \vspace{-2mm}
\resizebox{0.7\hsize}{!}{\begin{tabular}{c|c|c}
\hline
Model                              & Acc w/ Tie & Acc w/o Tie \\ \hline
RL w/o SFT                         & 0.387           & 0.513        \\
RL w/o $R_{\text{pref}}$          & 0.352           & 0.514        \\
REACT w/o DS & 0.519           & 0.725        \\
\textbf{REACT(\textit{Default})}                               & \textbf{0.610}           & \textbf{0.813}        \\ \hline
\end{tabular}}
\vspace{-2mm}
\label{tabel_ablation_score}
\end{table}
\begin{table}
  \centering
  \caption{\textbf{Ablation Study on RL Starting Point, SFT Epoch, and Loss Function in Distortion Recognition Task.} Our REACT model, trained with a two-stage paradigm (i.e., SFT and GRPO) and utilizing masked loss in the second epoch of SFT, achieves the best performance in distortion recognition.}
  \renewcommand{\arraystretch}{1.3}  
  \vspace{-2mm}
\resizebox{\hsize}{!}{\begin{tabular}{ccccccc}
\hline
\multicolumn{1}{c|}{\multirow{2}{*}{Ablations}}    & \multicolumn{3}{c|}{Distorted Frame}               & \multicolumn{3}{c}{Normal Frame} \\ \cline{2-7} 
\multicolumn{1}{c|}{}                              & Recall & Precision & \multicolumn{1}{c|}{F1-Score} & Recall  & Precision  & F1-Score  \\ \hline
\multicolumn{7}{c}{\textit{Ablation for SFT}}                                                                                                       \\ \hline
\multicolumn{1}{c|}{SFT \footnotesize{(\textbf{1 Epoch})}}                 & 0.399  & 0.920      & \multicolumn{1}{c|}{0.557}    & \underline{0.842}   & 0.235      & 0.367     \\
\multicolumn{1}{c|}{SFT \footnotesize{(\textbf{2 Epoch w/o ML})}} & 0.548  & \textbf{0.933}     & \multicolumn{1}{c|}{0.690}     & 0.797   & 0.254      & 0.385     \\
\multicolumn{1}{c|}{SFT \footnotesize{(\textbf{2 Epoch w/ ML})}}  & \underline{0.664}  & 0.899     & \multicolumn{1}{c|}{\underline{0.764}}    & 0.659   & \underline{0.300}        & \underline{0.413}    \\ \hline
\multicolumn{7}{c}{\textit{Ablation for RL}}                                                                                                        \\ \hline
\multicolumn{1}{c|}{RL w/o SFT}                    & 0.312  & \underline{0.924}     & \multicolumn{1}{c|}{0.467}    & \textbf{0.868}   & 0.196      & 0.319     \\
\multicolumn{1}{c|}{\textbf{Our REACT}}                          & \textbf{0.866}  & 0.825     & \multicolumn{1}{c|}{\textbf{0.845}}    & 0.594   & \textbf{0.771}      & \textbf{0.671}     \\ \hline
\end{tabular}}
\vspace{-6mm}
\label{tabel_ablation_attr}
\end{table}
To further assess the impact of each component and setting in our REACT model, we conduct a series of ablation studies on both human preference alignment in REACT-Video and distortion recognition in REACT-Frame. The results are presented in Table~\ref{tabel_ablation_score} and Table~\ref{tabel_ablation_attr}, respectively.

As shown in Table~\ref{tabel_ablation_score}, we explore the effects of RL starting point, reward design, and sampling mechanism on the human preference alignment task. Compared to the full REACT model, which effectively aligns with human preferences, the model trained directly from Qwen2.5-VL-7B without supervised fine-tuning (RL w/o SFT) shows a significant performance drop, with accuracies of 0.387 (w/ ties) and 0.513 (w/o ties). We attribute this decline to the difficulty of Qwen2.5-VL-7B in generating diverse scores, which limits the effectiveness of GRPO, as it heavily relies on the quality of rollout trajectories. This highlights the necessity of fine-tuning with pseudo-scores during the SFT stage. To further evaluate the impact of preference reward, we also conduct experiments with a binary reward model (RL w/o $R_{\text{pref}}$), where the reward is set to 0 or 1 based on whether the predicted preference matches the ground truth. As shown in Table~\ref{tabel_ablation_score}, omitting the preference reward significantly degrades performance, emphasizing its importance. Finally, comparing REACT with and without dynamic sampling (REACT w/o DS) reveals that the default configuration with dynamic sampling further enhances performance, thanks to its flexible sampling mechanism.

Table~\ref{tabel_ablation_attr} presents the results of the ablation study on RL starting point, SFT epoch, and loss function in the distortion recognition task. Similarly to the human preference alignment task, the model without supervised fine-tuning (RL w/o SFT) shows much lower performance, with an F1-score of 0.467 for distorted frames and 0.319 for normal frames, indicating its difficulty in recognizing structural distortions. When training starts from SFT for one or two epochs without masked loss, the F1-scores for distorted frames improve to 0.557 and 0.690, respectively. Performance continues to improve with the incorporation of masked loss in the second epoch, and the highest performance is achieved with the application of GRPO, underscoring the importance of these components in optimizing model performance.

\vspace{-3mm}
\section{Conclusion}
\vspace{-2mm}

In this work, we introduced REACT, a frame-level reward model specifically designed to evaluate structural distortions in generative videos. By integrating SFT and GRPO, REACT excels in recognizing and evaluating structural distortions, an aspect often overlooked by current SOTA video and image evaluators. Through extensive ablation studies and experiments on the REACT-Video and REACT-Frame benchmarks, we demonstrated that REACT outperforms existing models in both human preference alignment and distortion recognition tasks. This improvement stems from our detailed structural distortion taxonomy and the efficient CoT synthesis pipeline, which together provide a strong data foundation to enhance the ability of REACT to reason over video frames and detect structural distortions.

Future work will focus on extending reasoning capabilities of REACT beyond individual video frames to incorporate spatio-temporal semantics. This would enable the detection of issues like flash effects or sudden disappearances in generative videos, which require temporal information for accurate recognition, a problem that current video reward models have not yet addressed adequately.

\section*{Acknowledgements}
This work is supported by the by the National Science and Technology Major Project (2023ZD0121102).
{
    \small
    \bibliographystyle{ieeenat_fullname}
    \bibliography{main}
}

\appendix
\onecolumn
\clearpage
\setcounter{page}{1}
\maketitlesupplementary
\section{Detailed Taxonomy of Structural Distortion}
\label{taxonomy}
\begin{figure*}[!t]
\centering
\includegraphics[width=\textwidth]{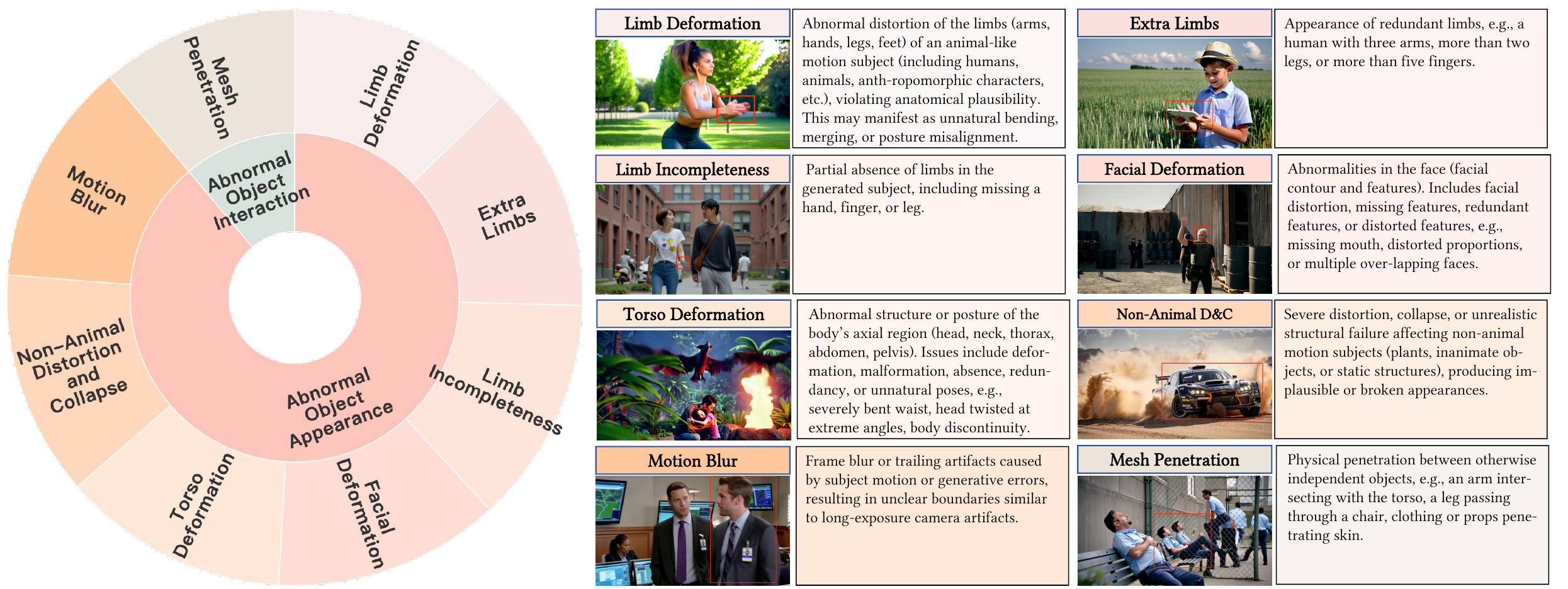}
\caption{\textbf{Detailed Explanation of Our Proposed Taxonomy of Structural Distortions in Generative Videos.} Representative examples for each distortion category are also provided.}
\label{fig:taxonomy}
\end{figure*}
Generative videos typically contain multiple interacting objects, therefore, we construct our taxonomy of structural distortions based on abnormalities in object appearance and object interaction within the video. We categorize structural distortions into two major groups: abnormal object appearance and abnormal object interaction. As illustrated in the section~\ref{subsec:data_prepara}, the former is further divided according to object characteristics into animal-centric, non-animal-centric, and motion-blur-related distortions. The animal-centric category includes limb deformation, extra limbs, limb incompleteness, torso deformation, and facial deformation. The non-animal-centric category corresponds to non-animal collapse and distortion. Abnormal object interaction primarily refers to mesh penetration. The complete taxonomy is illustrated in Fig.~\ref{fig:taxonomy}. Detailed definitions of each category are provided below:
\begin{itemize}
    \item \textbf{Limb Deformation:} Abnormal distortion of the limbs (arms, hands, legs, feet) of an animal-like subject (including humans, animals, anthropomorphic characters, etc.), violating anatomical plausibility. This may manifest as unnatural bending, merging, or posture misalignment, \eg, hyper-extended or reversed joints, twisted or fused fingers, abnormal stretching of arms, etc. In Fig.~\ref{fig:taxonomy}, the subject’s fingers are severely twisted and lose their normal shape and contour, which is representative of limb deformation.
    \item \textbf{Limb Incompleteness:} Partial absence of limbs in the generated subject, such as missing a hand, finger, or leg.
    \item \textbf{Extra Limbs:} The appearance of redundant limbs, \eg, a human with three arms, more than two legs, or more than five fingers. As shown in the second row and first column of Fig.~\ref{fig:taxonomy}, the woman displays anatomically implausible limb duplication, with no proper hands and only an arm remaining on her left side.
    \item \textbf{Torso Deformation:} Abnormal structure or posture of the body’s axial region (head, neck, thorax, abdomen, pelvis). Issues include deformation, malformation, absence, redundancy, or unnatural poses, \eg, severely bent waist, head twisted at extreme angles, body discontinuity. In Fig.~\ref{fig:taxonomy}, the woman’s head and back are positioned at an impossible angle, which can be categorized as torso deformation.
    \item \textbf{Facial Deformation:} Abnormalities in the face (facial contours and features). Includes facial distortion, missing features, redundant features, or distorted features, \eg, missing mouth, distorted proportions, or multiple overlapping faces. As shown in Fig.~\ref{fig:taxonomy}, the facial deformation refers to a distorted face that lacks normal anatomical structure and contour.
    \item \textbf{Mesh Penetration:} Physical penetration between otherwise independent objects, \eg, an arm intersecting with the torso, a leg passing through a chair, clothing or props penetrating the skin. As an example, two men sitting on a chair in Fig.~\ref{fig:taxonomy} appear to penetrate through the wire mesh, which is physically impossible.
    \item \textbf{Non-Animal Distortion and Collapse:} Severe distortion, collapse, or unrealistic structural failure affecting non-animal subjects (plants, inanimate objects, or static structures), producing implausible or broken appearances, such as the blurred and collapsed car front shown in the third row and second column of Fig.~\ref{fig:taxonomy}.
    \item \textbf{Motion Blur:} frame blur or trailing artifacts caused by subject motion or generative errors, resulting in unclear boundaries similar to long-exposure camera artifacts.
\end{itemize}
In addition to the above definitions, we further clarify the anatomical scope used throughout this taxonomy. The face includes both the facial contour and all facial features; limbs include arms, legs, hands, and feet; and the torso encompasses the head, neck, thorax, abdomen, and pelvis. For animals without limbs (\eg, snakes, fish) or stylized characters, all non-facial regions are considered part of the torso. Moreover, we do not treat abnormal posture as a standalone category. Instead, posture-related distortions affecting the axial region are classified as torso deformation, while posture anomalies occurring in the limbs fall under limb deformation.

\section{Dataset Annotation Rules}
\label{annotation}
To construct the annotated dataset that forms the foundation of our REACT framework, we collect a large-scale set of frame pairs following the procedure described in Section~\ref{subsec:data_prepara} and annotate them according to the taxonomy detailed in Appendix~\ref{taxonomy}. The annotation process comprises three components: (1) distortion recognition, (2) spatial grounding of each distortion label for every frame, and (3) human preference annotation, which we denote as GSB (\ie, Good / Same / Bad). Specifically, given a frame pair, annotators first examine each frame individually and assign bounding boxes corresponding to all annotated distortion types (\ie, attribution labels). They then determine a preference judgment for the pair based on the number and severity of the annotated bounding boxes and their associated attribution labels. To ensure consistency and reliability in evaluating structural distortions in generative videos, we establish detailed annotation guidelines for all three components.

For the distortion recognition task, annotators may assign at most three issue labels from the taxonomy to each frame. When a frame exhibits more than three issues, the selection is based primarily on the spatial extent and perceptual severity of the defects. For the grounding task, multiple bounding boxes may be assigned to a single attribution label when the corresponding distortion appears in multiple disjoint regions. Each bounding box must fully encompass the relevant distorted region such that the problematic content can be identified solely from information within the box, without relying on external context. When occlusion occurs, annotators approximate the full spatial extent of the affected area. In conclusion, bounding boxes should avoid unnecessary inclusion of irrelevant visual content to minimize interference from unrelated structures. For the human preference task, the frame containing fewer attribution labels and bounding boxes is preferred. A Same preference is assigned only when (1) both frames exhibit the same distortion types with comparable severity, or (2) neither frame contains identifiable structural distortion issues. Certain special cases follow additional principles outlined below:
\begin{itemize}
\item \textbf{Prioritizing Animal-Centric Labels.}
When more than three structural distortion types occur in a frame, animal-centric labels, textit{limb deformation, extra limbs, limb incompleteness, torso deformation, and facial deformation}, are prioritized. Non-animal collapse and distortion and mesh penetration follow, while \textit{motion blur} is considered last. This prioritization also applies to human preference annotation, where animal-centric distortions are treated as more severe in the GSB decision process.

\item \textbf{Distinguishing Motion Blur from Deformation and Collapse.}
Motion blur or trailing is annotated only when the subject displays explicit motion cues and retains an otherwise coherent and correct outline, with blurring localized around the moving edges. Blur, tearing, or deformation occurring in static objects (\eg, buildings, vegetation, background regions), \ie, non-animal entities under our taxonomy, is consistently attributed to non-animal collapse and distortion.

\item \textbf{Distinguishing Limb Incompleteness from Limb Deformation.}
Limb incompleteness is assigned when a limb component is entirely or partially absent, such as missing hands or feet, fewer than five fingers, or fully missing limbs. When a limb is present but structurally collapsed due to distortion, the appropriate label is limb deformation rather than limb incompleteness.
\end{itemize}

\section{Prompt Templates}
\label{prompt}
In this section, we provide a clear overview of the prompts used throughout the entire process. We first introduce the prompt designed for efficient CoT synthesis, as shown in Fig.~\ref{box:cot_instruction}. Specifically, we supply the annotated attribution labels together with their corresponding bounding boxes, and instruct Gemini to simulate the reasoning process that leads to these labels and bounding boxes. For structural distortion evaluation, we design two types of prompts based on our proposed taxonomy: one for the human preference alignment task and the other for the distortion recognition task. The prompt for human preference alignment is shown in Fig.~\ref{box:generation_instruction}, while the prompt for distortion recognition is presented in Fig.~\ref{box:recognition}. By incorporating detailed explanations of each distortion category, these prompts enable REACT to develop a more comprehensive understanding of structural distortions in generative videos, thereby producing more accurate evaluation results.
\begin{figure*}[h]
\begin{tcolorbox}[
    breakable,
    width=\textwidth,
    colback=white,
    colframe=black,
    enhanced,
    sharp corners,
    boxrule=1pt,
    drop shadow,
    coltitle=white,
    title=\textbf{Text Prompt for Our REACT in Human Preference Alignment Task},
    valign=center
]

\small

\textbf{What is your overall rating on the visual quality of this frame?} The rating should be a floating-point number between 1 and 5, rounded to two decimal places. A rating of 1 represents very poor visual quality, and a rating of 5 represents excellent visual quality. The visual quality issues to be considered include the following:

\begin{itemize}[leftmargin=*]
    \item \textbf{Limb Deformation:} Abnormal distortion of the limbs (arms, hands, legs, feet) of an animal-like subject (including humans, animals, anthropomorphic characters, etc.), violating anatomical plausibility. This may manifest as unnatural bending, merging, or posture misalignment, e.g., hyper-extended or reversed joints, twisted or fused fingers, abnormal stretching of arms, etc.
    \item \textbf{Limb Incompleteness:} Partial absence of limbs in the generated subject, such as missing a hand, finger, or leg.
    \item \textbf{Extra Limbs:} The appearance of redundant limbs, e.g., a human with three arms, more than two legs, or more than five fingers.
    \item \textbf{Torso Deformation:} Abnormal structure or posture of the body’s axial region (head, neck, thorax, abdomen, pelvis). Issues include deformation, malformation, absence, redundancy, or unnatural poses, e.g., severely bent waist, head twisted at extreme angles, body discontinuity.
    \item \textbf{Facial Deformation:} Abnormalities in the face (facial contours and features). Includes facial distortion, missing features, redundant features, or distorted features, e.g., missing mouth, distorted proportions, or multiple overlapping faces.
    \item \textbf{Mesh Penetration:} Physical penetration between otherwise independent objects, e.g., an arm intersecting with the torso, a leg passing through a chair, clothing or props penetrating the skin.
    \item \textbf{Non-Animal Distortion and Collapse:} Severe distortion, collapse, or unrealistic structural failure affecting non-animal subjects (plants, inanimate objects, or static structures), producing implausible or broken appearances.
    \item \textbf{Motion Blur:} frame blur or trailing artifacts caused by subject motion or generative errors, resulting in unclear boundaries similar to long-exposure camera artifacts.
\end{itemize}

Please first assess whether the frame exhibits any of the issues listed above, and then provide an overall rating for the picture. The final answer should be returned in JSON format with the following keys:
\vspace{-0.5mm}
\begin{tcolorbox}[colback=gray!5, colframe=gray!30, boxrule=0.3pt, left=2mm, right=2mm, top=0.5mm, bottom=0.5mm, listing only]
\scriptsize
\begin{verbatim}
{
    "Attribution labels": [A list of the detected issues or "null" if none are found],
    "rating": [The score]
}
\end{verbatim}
\end{tcolorbox}
\end{tcolorbox}
\end{figure*}

\begin{figure*}[ht]
\centering
\caption{\textbf{Text Prompt for Our REACT in Human Preference Alignment Task.}}
\label{box:generation_instruction}
\end{figure*}
\begin{figure*}
\begin{tcolorbox}[
    breakable,
    width=\textwidth,
    colback=white,
    colframe=black,
    enhanced,
    sharp corners,
    boxrule=1pt,
    drop shadow,
    coltitle=white,
    title=\textbf{Text Prompt for Our REACT in Distortion Recognition Task},
    valign=center
]

\small

Analyze the provided frame to determine whether it exhibits any of the following visual quality issues: Limb Deformation, Torso Deformation, Facial Deformation, Limb Incompleteness, Extra Limbs, Mesh Penetration, Non-animal Distortion and Collapse, Motion Blur.

\begin{itemize}[leftmargin=*]
    \item \textbf{Limb Deformation:} Abnormal distortion of the limbs (arms, hands, legs, feet) of an animal-like subject (including humans, animals, anthropomorphic characters, etc.), violating anatomical plausibility. This may manifest as unnatural bending, merging, or posture misalignment, e.g., hyper-extended or reversed joints, twisted or fused fingers, abnormal stretching of arms, etc.
    \item \textbf{Limb Incompleteness:} Partial absence of limbs in the generated subject, such as missing a hand, finger, or leg.
    \item \textbf{Extra Limbs:} The appearance of redundant limbs, e.g., a human with three arms, more than two legs, or more than five fingers.
    \item \textbf{Torso Deformation:} Abnormal structure or posture of the body’s axial region (head, neck, thorax, abdomen, pelvis). Issues include deformation, malformation, absence, redundancy, or unnatural poses, e.g., severely bent waist, head twisted at extreme angles, body discontinuity.
    \item \textbf{Facial Deformation:} Abnormalities in the face (facial contours and features). Includes facial distortion, missing features, redundant features, or distorted features, e.g., missing mouth, distorted proportions, or multiple overlapping faces.
    \item \textbf{Mesh Penetration:} Physical penetration between otherwise independent objects, e.g., an arm intersecting with the torso, a leg passing through a chair, clothing or props penetrating the skin.
    \item \textbf{Non-Animal Distortion and Collapse:} Severe distortion, collapse, or unrealistic structural failure affecting non-animal subjects (plants, inanimate objects, or static structures), producing implausible or broken appearances.
    \item \textbf{Motion Blur:} frame blur or trailing artifacts caused by subject motion or generative errors, resulting in unclear boundaries similar to long-exposure camera artifacts.
\end{itemize}

If any issues are detected, identify the three most severe ones. Return the result in JSON format with the following keys:
\vspace{-0.5mm}
\begin{tcolorbox}[colback=gray!5, colframe=gray!30, boxrule=0.3pt, left=2mm, right=2mm, top=0.5mm, bottom=0.5mm, listing only]
\scriptsize
\begin{verbatim}
{
    "Attribution labels": [A list of the detected issues or "null" if none are found]
}
\end{verbatim}
\end{tcolorbox}
\end{tcolorbox}
\end{figure*}

\begin{figure*}[h]
\centering
\caption{\textbf{Text Prompt for Our REACT in Distortion Recognition Task.}}
\label{box:recognition}
\end{figure*}
\section{Additional Experiments Results}
\label{add_exp}
\subsection{Evaluation Prompt}
When evaluating human preference alignment with REACT-Video, we apply each video reward model, VideoScore2, UnifiedReward, and VideoReward, using their original prompts, which are designed to assess multiple aspects of video quality holistically. For general MLLMs, we adopt the same prompt used in REACT, which includes detailed descriptions of each distortion type and the principles for assigning point-wise scores. This prompt guides the models to generate distortion-aware point-wise quality assessments.For image evaluators, we use their native prompts and further introduce the REACT prompt as a refined supplementary prompt, allowing these models to incorporate auxiliary knowledge about structural distortions in generative videos during the additional experiments.

When evaluating distortion recognition with REACT-Frame, only image evaluators and general MLLMs are responsible for this task. All models, including MagicAccessor, are instructed using the prompt shown in Fig.~\ref{box:recognition}, which contains detailed explanations of all attribution labels associated with structural distortion. This is because all these models are trained or adapted from general-purpose MLLMs capable of instruction following, enabling them to perform the required annotation tasks under a well-specified prompt.

\subsection{Evaluation Metrics}
For the human preference alignment evaluation, we use preference accuracy as the metric to assess the performance of REACT. Specifically, we report accuracy with tie and without tie. Accuracy without tie directly compares the point-wise scores of the two frames in each pair and assigns the preference to the frame with the higher score. For accuracy with tie, we additionally consider the cases where the two frames are essentially equivalent, that is, if the score difference between the two frames falls below a predefined threshold, the pair is treated as a tie. Since all baselines are prompted to produce point-wise scores rather than explicitly comparing the frame pairs, we first convert their point-wise scores into pairwise preferences following the above procedure. As described in Section~\ref{exp:main_results}, we compute the VQ score and MQ score, and their combined overall score, to derive the final preference for video evaluators. For VideoReward, VQ and MQ correspond to the ``visual quality'' and ``motion quality'' dimensions, respectively.
For VideoScore, VQ corresponds to ``visual quality'' and MQ corresponds to ``physical/common-sense consistency''.
For UnifiedReward, VQ maps to ``visual quality,'' while MQ is defined as the average of ``temporal consistency'' and ``factual consistency''. 

For the distortion recognition task, we evaluate the performance of REACT using precision, recall, and F1-score, which measure how accurately the model identifies frames suffering from structural distortions. The calculation is defined as follows:
\begin{align}
&\text{Precision} = \frac{TP}{TP + FP}, \\
&\text{Recall} = \frac{TP}{TP + FN}, \\
&\text{F1-score} = \frac{2 \cdot \text{Precision} \cdot \text{Recall}}
{\text{Precision} + \text{Recall}},
\end{align}
where TP, FP, and FN denote the number of true positives, false positives, and false negatives, respectively. Precision reflects the accuracy of positive predictions, i.e., the proportion of predicted positive samples that are truly positive. Recall reflects the coverage of the model, i.e., the proportion of true positive samples that are correctly identified. F1-score provides a comprehensive measure of overall performance by balancing precision and recall.
\subsection{Additional Human Preference Alignment}
\begin{table}
  \centering
  \caption{\textbf{Additional Experiments on GenAI Benchmark and VideoGen-RewardBench.}}
  \renewcommand{\arraystretch}{1.3}  
\resizebox{\hsize}{!}{\begin{tabular}{c|cccc|cc}
\hline
                        & \multicolumn{4}{c|}{VideoGen-RewardBench}                                                                                                                                                & \multicolumn{2}{c}{}                                                       \\ \cline{2-5}
                        & \multicolumn{2}{c|}{VQ}                                                                         & \multicolumn{2}{c|}{MQ}                                                    & \multicolumn{2}{c}{\multirow{-2}{*}{GenAI}}                                \\ \cline{2-7} 
\multirow{-3}{*}{Model} & { Acc w/ Tie} & \multicolumn{1}{c|}{Acc w/o Tie}    & { Acc w/ Tie} & Acc w/o Tie    & {Acc w/ Tie} & Acc w/o Tie    \\ \hline
VideoScore2             & 0.424                                                     & \multicolumn{1}{c|}{0.515}          & 0.383                                                     & 0.706          & 0.391                                                     & 0.616          \\
UnifiedReward           & {\underline{ 0.589}}                                               & \multicolumn{1}{c|}{{\underline {0.701}}}    & {\underline {0.475}}                                               & {\underline {0.749}}    & \textbf{0.548}                                            & {\underline {0.709}}    \\
VideoReward             & \textbf{0.660}                                            & \multicolumn{1}{c|}{\textbf{0.746}} & \textbf{0.596}                                            & \textbf{0.756} & {\underline{ 0.491}}                                               & \textbf{0.728} \\
Qinsight                & 0.367                                                     & \multicolumn{1}{c|}{0.533}          & 0.372                                                     & 0.663          & 0.376                                                     & 0.571          \\
Our REACT               & 0.402                                                     & \multicolumn{1}{c|}{0.538}          & 0.386                                                     & 0.626          & 0.376                                                     & 0.581          \\ \hline
\end{tabular}}
\vspace{-4mm}
\label{tabel_additional}
\end{table} \label{add_exp: align}
We also conduct experiments on the GenAI benchmark and VideoGen-RewardBench. The former is a reward benchmark for generative models, annotated with human preferences over visual content produced by image editing, image generation, and video generation models. We use the subset corresponding to generative video to evaluate the performance of our REACT on video quality assessment. The latter benchmark extends VideoGen-Eval to construct a human-preference dataset for evaluating reward models on modern text-to-video (T2V) models. As shown in Table~\ref{tabel_additional}, REACT is slightly inferior to video-based evaluators in terms of overall preference accuracy. We attribute this to the fact that REACT is grounded in a new preference formulation that emphasizes structural distortions—an aspect not explicitly modeled in existing video evaluation methods. Nevertheless, REACT outperforms the image-based evaluator Q-Insight, demonstrating its stronger ability to assess generative video quality.
\begin{table*}[]
\caption{\textbf{Comparison of Reward Models for Improving Video Generation Quality on VBench.} Our REACT substantially improves video generation quality, and integrating it with other SOTA reward models yields additional gains.}
\resizebox{\textwidth}{!}{\begin{tabular}{cccccc}
\hline
\multicolumn{1}{c|}{\multirow{2}{*}{Model}} & \multicolumn{5}{c}{VBench}                                                                                                \\ \cline{2-6} 
\multicolumn{1}{c|}{}                       & \multicolumn{1}{c|}{Background Consistency $\uparrow$}    & \multicolumn{1}{c|}{Dynamic Degree $\uparrow$}    & \multicolumn{1}{c|}{Imaging Quality $\uparrow$}    & \multicolumn{1}{c|}{Subject Consistency $\uparrow$}    & Aesthetic Quality  $\uparrow$   \\ \hline
\multicolumn{1}{c|}{Wan-2.1-1.3B}               & \multicolumn{1}{c|}{0.951} & \multicolumn{1}{c|}{0.527} & \multicolumn{1}{c|}{0.649} & \multicolumn{1}{c|}{0.948} & 0.522 \\ \hline
\multicolumn{6}{c}{\textit{w/ Best-of-N}}                                                                                                                                        \\ \hline
\multicolumn{1}{c|}{UnifiedReward (UR)}     & \multicolumn{1}{c|}{0.957} & \multicolumn{1}{c|}{0.541} & \multicolumn{1}{c|}{0.674} & \multicolumn{1}{c|}{0.959} & 0.542 \\
\multicolumn{1}{c|}{REACT}                  & \multicolumn{1}{c|}{0.955} & \multicolumn{1}{c|}{0.527} & \multicolumn{1}{c|}{0.675} & \multicolumn{1}{c|}{0.955} & 0.547 \\
\multicolumn{1}{c|}{UR+REACT}               & \multicolumn{1}{c|}{0.957} & \multicolumn{1}{c|}{0.541} & \multicolumn{1}{c|}{0.675} & \multicolumn{1}{c|}{0.960} & 0.547 \\ \hline
\multicolumn{6}{c}{\textit{w/ Flow-DPO}}                                                                                                                                         \\ \hline
\multicolumn{1}{c|}{UnifiedReward (UR)}     & \multicolumn{1}{c|}{0.971} & \multicolumn{1}{c|}{0.542} & \multicolumn{1}{c|}{0.690} & \multicolumn{1}{c|}{0.977} & 0.547 \\
\multicolumn{1}{c|}{REACT}                  & \multicolumn{1}{c|}{0.963} & \multicolumn{1}{c|}{0.536} & \multicolumn{1}{c|}{0.691} & \multicolumn{1}{c|}{0.977} & 0.549 \\
\multicolumn{1}{c|}{UR+REACT}               & \multicolumn{1}{c|}{\textbf{0.981}} & \multicolumn{1}{c|}{\textbf{0.554}} & \multicolumn{1}{c|}{\textbf{0.694}} & \multicolumn{1}{c|}{\textbf{0.998}} & \textbf{0.550} \\ \hline
\end{tabular}}
\label{tab:main_rl}
\end{table*}

\subsection{Performance on Improving Video Generation}
\label{add_exp: rl}
To further demonstrate the effectiveness of REACT in improving the visual quality of generated videos, we integrate it into two representative paradigms, Best-of-$N$ sampling and Flow-DPO \cite{liu2025improving}, on the open-source video generation model Wan-2.1-1.3B \cite{wan2025wan}, and compare it against state-of-the-art reward models on VBench\cite{huang2024vbench}. For Best-of-$N$ sampling, we generate five videos for each prompt and select the one with the highest reward score. For Flow-DPO, we sample 5.7K prompts from the training dataset and generate videos with Wan-2.1-1.3B, where the positive and negative samples are determined according to the reward scores assigned by the corresponding reward model.

As shown in Tab.~\ref{tab:main_rl}, under Best-of-$N$ sampling, REACT alone achieves performance competitive with UnifiedReward, slightly outperforming it in Imaging Quality and Aesthetic Quality while maintaining comparable results on Background Consistency and Subject Consistency. These results indicate that REACT can effectively improve the visual fidelity of generated videos. Under Flow-DPO post-training, REACT further surpasses UnifiedReward in Imaging Quality and Aesthetic Quality, demonstrating that accurate assessment of structural distortions provides a more reliable supervision signal for video generation

Furthermore, we evaluate a simple reward fusion strategy that combines REACT and UnifiedReward by averaging their scores as the final reward for generated videos. This combined model yields additional gains in both paradigms and achieves the best performance across all evaluated metrics. These results suggest that REACT captures structural cues that are complementary to existing reward models, and that incorporating such feedback can further improve overall video generation quality.

\subsection{Case Study}
\begin{figure*}[!t]
\centering
\includegraphics[width=\textwidth]{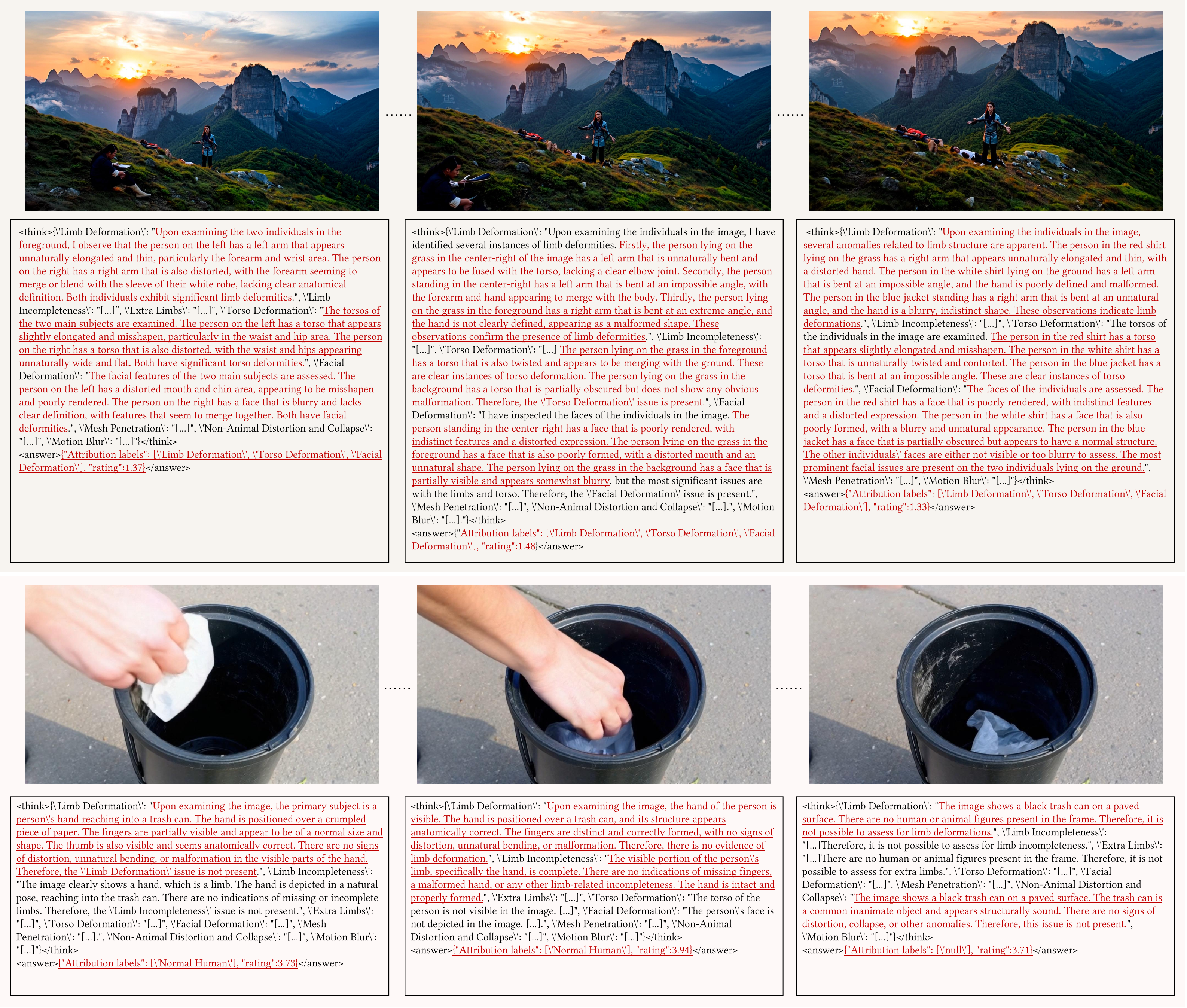}
\caption{\textbf{Case Study of REACT for Distortion Evaluation in Generative Videos.} The two presented video cases illustrate that REACT effectively identifies structural distortions and produces reliable point-wise assessments for generative videos.}
\label{fig:case}
\end{figure*}
We present qualitative results in Fig.~\ref{fig:case}. In the first row, the video contains severe structural distortions, and our REACT successfully identifies all distortions and assigns a reliable point-wise score reflective of its low visual quality. In contrast, the second row shows a high-quality video without structural distortions. Likewise, REACT correctly recognizes it as a normal video and provides a correspondingly high score. These qualitative examples clearly demonstrate that REACT performs well in distortion evaluation, both in accurately recognizing structural distortions and in assigning reliable point-wise scores.

\onecolumn
\begin{tcolorbox}[
    breakable,
    width=\textwidth,
    colback=white,
    colframe=black,
    enhanced,
    sharp corners,
    boxrule=1pt,
    drop shadow,
    coltitle=white,
    title=\textbf{Text Prompt for CoT Synthesis},
    valign=center
]
\small

{\large \textbf{Role and Goal}} \\
You are an expert in generated frame quality assessment. \\
You are given an frame that may have \textbf{dynamic quality issues}, \textbf{along with a set of annotations \texttt{"<label with bbox>"} (each item pairs an attribution \texttt{"<label>"} with a corresponding bounding box \texttt{"<bbox>"})}.

\textbf{Annotation definitions:}

\begin{itemize}[leftmargin=*]
    \item \textbf{\texttt{<label>}:} List[choice] 
    
    Each entry denotes a dynamic quality issue present in the frame. Candidate labels include: \textbf{\textit{limb deformation, limb incompleteness, extra limbs, torso deformation, facial deformation, mesh penetration, non-animal distortion and collapse, motion blur, and no issue.}}
    \item \textbf{\texttt{<bbox>}:} List[list] 
    
    Each entry is $[x_1, y_1, x_2, y_2]$, where:(1)~\textbf{$x_1$:} x-coordinate of the top-left corner of the bounding box;(2)~\textbf{$y_1$:} y-coordinate of the top-left corner of the bounding box;(3)~\textbf{$x_2$:} x-coordinate of the bottom-right corner of the bounding box;(4)~\textbf{$y_2$:} y-coordinate of the bottom-right corner of the bounding box.
    
    \item \textbf{\texttt{<label with bbox>}:} List[tuple] 
    
    Each item consists of an attribution label $<\text{label}>$ and its corresponding bounding box $<\text{bbox}>$.
\end{itemize}

\textbf{Task description:} 

Your task is: \textbf{Assume you don’t know the content of these labels. Based only on visual features you observe in the frame, analyze step by step what problems are present}, and ultimately infer the phenomenon corresponding to the attribution label. Bounding box information serves only as a localization reference to help you confirm the problematic area, but it must not drive your judgment. Consider the frame holistically, proceed step by step, and naturally infer the likely attribution label. This process must reflect a professional \textbf{Chain of Thought}.

\textbf{Output requirement:}

The final result must be returned as a \textbf{complete JSON file}. \textbf{Do not output any content or explanatory text outside the JSON.}\\ 

{\large \textbf{Core Instructions}}

\begin{enumerate}[leftmargin=10pt]
    \item \textbf{Chain of Thought (CoT)}
    
    Generate the analysis process corresponding to each label based on the frame itself, meeting the following requirements:
    \begin{itemize}[leftmargin=10pt]
        \item Show a typical, professional analysis workflow for generated frame quality assessment, determining whether the frame exhibits any of the following issues: \textit{limb deformation, limb incompleteness, extra limbs, torso deformation, facial deformation, mesh penetration, non-animal distortion and collapse, motion blur, and no issue. Details:}
        \begin{itemize}[leftmargin=10pt]
            \item \textbf{\texttt{Limb Deformation}}: Abnormal distortion of the limbs (arms, hands, legs, feet) of an animal-like motion subject (including humans, animals, anthropomorphic characters, etc.), violating anatomical plausibility. This may manifest as unnatural bending, merging, or posture misalignment, e.g., hyper-extended or reversed joints, twisted or fused fingers, abnormal stretching of arms, etc.
            \item \textbf{\texttt{Limb Incompleteness}}: Partial absence of limbs in the generated subject, e.g., missing a hand, finger, or leg.
            \item \textbf{\texttt{Extra Limbs}}: Appearance of redundant limbs, e.g., a human with three arms, more than two legs, or more than five fingers.
            \item \textbf{\texttt{Torso Deformation}}: Abnormal structure or posture of the body’s axial region (head, neck, thorax, abdomen, pelvis). Issues include deformation, malformation, absence, redundancy, or unnatural poses, e.g., severely bent waist, head twisted at extreme angles, body discontinuity.
            \item \textbf{\texttt{Facial Deformation}}: Abnormalities in the face (facial contour and features). Includes facial distortion, missing features, redundant features, or distorted features, e.g., missing mouth, distorted proportions, or multiple overlapping faces.
            \item \textbf{\texttt{Mesh Penetration}}: Physical penetration between otherwise independent objects, e.g., an arm intersecting with the torso, a leg passing through a chair, clothing or props penetrating skin.
            \item \textbf{\texttt{Non-animal Distortion and Collapse}}: Severe distortion, collapse, or unrealistic structural failure affecting non-animal motion subjects (plants, inanimate objects, or static structures), producing implausible or broken appearances.
            \item \textbf{\texttt{Motion Blur}}: frame blur or trailing artifacts caused by subject motion or generative errors, resulting in unclear boundaries similar to long-exposure camera artifacts.
            \item \textbf{\texttt{No Issue}}: The frame has no apparent dynamic quality defects overall.
        \end{itemize}
        \item \textbf{Source of evidence}: Base reasoning and judgments \textbf{only on observable visual features}.
        \item \textbf{Independence constraint}: \textbf{Do not use the attribution labels or their bounding boxes for reverse validation or inference}; they may be used only for comparison after your reasoning is complete.
        \item \textbf{Factuality}: \textbf{Do not fabricate elements that are not present in the frame} (e.g., inventing objects/people/actions).
        \item You may naturally arrive at the attribution indicated by the labels, but the process must be based on observation rather than hints from labels.
        \item \textbf{Analyze each attribution label one by one, with an independent chain of thought for each label.}
    \end{itemize}
    \textbf{Consistency requirement}: The final inferred attribution must match the ground-truth \texttt{"<label>"}, and the problematic regions indicated during reasoning must strictly align with \texttt{"<label with bbox>"}.

    \item \textbf{JSON Output Format} 
    
    Your output must be a \textbf{clear, syntactically correct, valid JSON object} where each attribution label is a \textbf{key}, and the corresponding analysis process is the \textbf{value}. \textbf{Do not output anything outside the JSON structure. The return must be valid JSON; Markdown styling or pseudo-JSON is strictly forbidden.}

    \textbf{JSON format:}
\begin{tcolorbox}[colback=gray!5, colframe=gray!30, boxrule=0.3pt, left=2mm, right=2mm, top=0.5mm, bottom=0.5mm, listing only]
\scriptsize
\begin{lstlisting}[basicstyle=\ttfamily,breaklines=true]
{
  "COT": {
    "Limb Deformation": "The reasoning process determining whether this issue exists in the frame",
    "Limb Incompleteness": "The reasoning process determining whether this issue exists in the frame",
    "Extra Limbs": "The reasoning process determining whether this issue exists in the frame",
    "Torso Deformation": "The reasoning process determining whether this issue exists in the frame",
    "Facial Deformation": "The reasoning process determining whether this issue exists in the frame",
    "mesh penetration": "The reasoning process determining whether this issue exists in the frame",
    "non-animal distortion and collapse": "The reasoning process determining whether this issue exists in the frame",
    "Motion Blur": "The reasoning process determining whether this issue exists in the frame"
  },
  "Attribution Label": "Based on the CoT, the label corresponding to the issue that truly exists in the frame",
  "Problem Region": "Based on the CoT, the region corresponding to the issue that truly exists in the frame"
}
\end{lstlisting}
\end{tcolorbox}

    \textbf{Field descriptions}

\begin{itemize}[leftmargin=10pt]
    \item \textbf{COT}: \textbf{[To fill]} For the given frame, \textbf{analyze and verify each issue label (limb deformation, limb incompleteness, extra limbs, torso deformation, facial deformation, mesh penetration, non-animal distortion and collapse, motion blur) in turn}, determining whether the issue exists and writing out the complete reasoning process for each label in order. If none of these issues appear, you may provide the **“no Issue”** attribution label. Suggested content includes:
    \begin{itemize}[leftmargin=10pt]
        \item \textbf{Input evidence} (data source, frame/region/timestamp, visible features);
        \item \textbf{Reasoning steps} (logical transition from evidence to decision and exclusion tests; note: the visual evidence used in reasoning should \textbf{fall within the region indicated by \texttt{"<label with bbox>"}});
        \item \textbf{Conclusion} (final judgment).
    \end{itemize}
    \item \textbf{Attribution Label}: \textbf{[To fill]} The anomaly category inferred from the CoT analysis, \textbf{which must strictly match the ground-truth \texttt{"<label>"}}.
    \item \textbf{Problem Region}: \textbf{[To fill]} The frame region corresponding to the inferred attribution. \textbf{Each anomalous region must match the meaning of the attribution label, and the overall region must strictly match the ground-truth bounding box \texttt{"<label with bbox>"}}.
\end{itemize}
\end{enumerate}

{\large \textbf{Please begin your analysis.}}

\end{tcolorbox}

\begin{figure*}[h]
\centering
\vspace{-1mm}
\caption{\textbf{Text prompt for Efficient CoT Synthesis.}}
\vspace{-6mm}
\label{box:cot_instruction}
\end{figure*}

\end{document}